\documentclass[10pt,twocolumn,letterpaper]{article}

\usepackage{iccv}
\usepackage{times}
\usepackage{epsfig}
\usepackage{graphicx}
\usepackage{amsmath}
\usepackage{amssymb}
\usepackage{textcomp} 
\usepackage{multirow}
\usepackage{algorithm}
\usepackage[noend]{algpseudocode}
\DeclareMathOperator*{\argmax}{arg\,max}

\DeclareMathOperator*{\round}{round}
\makeatletter
\def\BState{\State\hskip-\ALG@thistlm}
\makeatother

\newcommand{\code}{\texttt} 

\usepackage[breaklinks=true,bookmarks=false]{hyperref}

\iccvfinalcopy 

\def\iccvPaperID{450} 

\ificcvfinal\fi

\begin{document}

\title{DSConv: Efficient Convolution Operator}

\author{Marcelo Gennari do Nascimento\\
University of Oxford\\
Active Vision Lab\\
{\tt\small marcelo@robots.ox.ac.uk}
\and
Roger Fawcett\\
Intel Corporation\\
https://www.omnitek.tv/about\\
{\tt\small roger.fawcett@intel.com}
\and
Victor Adrian Prisacariu\\
University of Oxford\\
Active Vision Lab\\
{\tt\small victor@robots.ox.ac.uk}
}

\maketitle
\ificcvfinal\thispagestyle{empty}\fi

\begin{abstract}
Quantization is a popular way of increasing the speed and lowering the memory usage of Convolution Neural Networks (CNNs). When labelled training data is available, network weights and activations have successfully been quantized down to 1-bit. The same cannot be said about the scenario when labelled training data is not available, \eg when quantizing a pre-trained model, where current approaches show, at best, no loss of accuracy at 8-bit quantizations.

We introduce \textit{DSConv}, a flexible quantized convolution operator that replaces single-precision operations with their far less expensive integer counterparts, while maintaining the probability distributions over both the kernel weights and the outputs. We test our model as a plug-and-play replacement for standard convolution on most popular neural network architectures, ResNet, DenseNet, GoogLeNet, AlexNet and VGG-Net and demonstrate state-of-the-art results, with less than 1\% loss of accuracy, without retraining, using only 4-bit quantization. We also show how a distillation-based adaptation stage with unlabelled data can improve results even further.
\end{abstract}

\section{Introduction}
A popular method to make neural networks faster and use less memory is quantization, which replaces 32-bit floating point weights and, potentially, activations with lower bit (\ie lower precision) representations, while aiming to maintain accuracy. 

Quantization is often used in neural network \textit{compression}. This aims to reduce the memory occupied by the network weights as much as possible to, for example, lower the overall memory footprint required to store the network. It can also be used to increase neural network inference speed (\textit{fast inference}), when applied to both weights and activations, by substituting expensive floating-point Multiply and Accumulate (MAC) operations with cheaper alternatives such as integer, bitwise operations or addition-only operations. 

The best quantization results are achieved when labelled training data is available, as the quantized model can be fitted to the dataset, which feeds the training algorithm with prior knowledge of what the activation maps will look like and what the expected output will be. Maintaining a high accuracy becomes much more difficult when only a pre-trained model is available.

In this paper we focus on this latter scenario, and quantize both weights and activations to produce neural networks that are both smaller and have faster inference. Our key insight is that, in the absence of training data, this can be best achieved by forcing the probability distributions over the weights and activations of the low precision quantized model to mirror those of the original full-precision model. We introduce a novel convolution operator, which we call \textit{DSConv}, that factorises the convolution weights into (i) a low-precision component with the same size as the original kernel and (ii) a high-precision distribution shift component, with a variable size (\eg as small as one \textit{float 32} value per kernel). A similar procedure, inspired by the block floating point approach \cite{Wilkinson:1994:REA:1096474}, is used to quantize activations. We also show that accuracy can be improved when using a distillation \cite{hinton2015distilling} inspired weight adaptation approach, that uses the original pre-trained model and unlabelled input data. 

The main contribution of this paper is the introduction of a convolution operator that (i) serves as a ``drop-in and play" replacement for standard convolution and uses low-bit fixed point computation for the bulk of operations without the need of retraining using labelled data, and (ii) provides a hyperparameter that can be tuned to favor accuracy or memory usage/speed of computation for any given task. Our quantization strategy is able to achieve state-of-the-art results, as demonstrated by our experimental section.

The remainder of this paper is structured as follows. \S\ref{sec:relatedWork} presents the previous papers on quantization. \S\ref{sec:method} explains the method in detail. \S\ref{sec:experiments} shows the results of the experiments performed in a variety of architectures and settings. \S\ref{sec:conclusion} concludes the paper with a discussion on its performance and possible applications and limitations.

\section{Related Work}
\label{sec:relatedWork}
The use of low-bit width networks saves a significant amount of memory and computation, especially when targeted to custom hardware. For example, for 8-bit operations, \cite{krishnamoorthi2018quantizing} reports up to 10x increase in speed, and \cite{highperformance} reports up to 30x in energy saving and chip area. We categorize the previous research that tries to increase the neural network efficiency in two groups:

\noindent\textbf{Quantization with labelled data.} Most research in neural network quantization has focused on problems that involve retraining, by either starting from scratch or by adapting from an existing pre-trained network. BinaryConnect, BWN and TWN \cite{Courbariaux2015BinaryConnectTD, Rastegari_2016, li2016ternary} use 1-bit and ternary weights to make the FP-MACS addition only. XNOR-Net and BNN \cite{Rastegari_2016, Courbariaux2015BinaryConnectTD} applied 1-bit quantized weight and activations to ImageNet for fast inference, at the cost of a significant drop in accuracy. WRPN \cite{mishra2017wrpn} improved this accuracy by using wider versions of the same architectures. Early demonstrations of low-bit network acceleration in custom hardware include Ristretto \cite{gysel2018ristretto}, which also uses data to quantize the network to 8-bit models. Many other papers followed, by also training the quantization scheme, using binary basis vectors, such as in LQ-Nets \cite{Zhang_2018}, Halfway Gaussian Quantization (HWGQ) \cite{Cai_2017}, and ABC-Net \cite{lin2017towards}. DoReFa-Net \cite{zhou2016dorefa} also quantized gradients, alongside weights and activations.

The compression problem has also mostly been dealt with by using retraining with access to labelled data. In DeepCompression \cite{han2015deep}, two of the three steps of the algorithm require retraining (pruning and quantization), with Huffman Encoding being performed without the need for data. HashedNet \cite{chen2015compressing} use the ``hashing trick" to save significant amounts of memory when storing the network, but still require labelled data for tuning. The more recent approach \cite{polino2018model}, uses distillation \cite{NIPS2014_5484, hinton2015distilling} obtain compressed weights, but also requires the full labelled training set.

Several approaches introduce novel float data formats. Examples are Dynamic Fixed Point \cite{courbariaux2014training}, which substitutes the normal floating point numbers with a mix of both fixed and floating point; and Flexpoint \cite{koster2017flexpoint}, which aims to leverage the range of floating point numbers and the computational complexity of fixed point and promises to perform forward and backwards operations with limited range. The idea of substituting the representation of single-precision (FP32) values in favour of other formats is also adopted in the \textit{bfloat16} format in Tensorflow \cite{tensorflow2015-whitepaper}, which employs the \textit{binary float 16} format that uses 7-bits for the mantissa instead of the usual 23-bits. 

\noindent\textbf{Quantization without labelled data.} Whereas the problem of quantizing with labelled data has been researched extensively, the problem of quantizing without data has received far less attention. Recent papers that explore this possibility are \cite{zhao2019improving, choukroun2019lowbit, jacob2018quantization, banner2018posttraining}, which either report results only for 8-bit quantization or employ calibration data of some sort - \ie an unlabelled small fraction of the validation dataset that is used for weight adaptation. Industry approaches have implemented quantization techniques that use only a small amount of unlabelled data, in systems such as  TensorRT \cite{TensorRT}. In this instance, they can successfully quantize a network to 8-bits with no loss of accuracy (sometimes even with improved accuracy) from 1000s of sampled images \cite{TensorRT}. Other examples include the Google TPU, and Project Brainwave (\cite{fowers2018configurable, 8344479}), all of which quantize neural networks to 8-bits for fast inference. Another work that shows that 8-bit quantization does not affect efficiency significantly is \cite{krishnamoorthi2018quantizing}, where they show that this is true even when quantizing both activation and weights.

In this paper, we show that quantization can be done effectively to 4-bits for both weights and activations, without the need of retraining labelled data, with further potential improvements when using adaptation with unlabelled data. 

\section{Method}
\label{sec:method}
For a given neural network inference $f(x)$, the prediction of $f(Mx)$ should be identical (considering that biases are scaled accordingly), for $M \in \mathbb{R}$. \textit{DSConv} is built on the intuition that this property holds for some nonlinear transforms of $x$, as long as the relative distribution of the weights and activation values remains the same. We believe one such transform to be quantization, \ie we can scale and bias quantized weights and activations in a way that is friendly for low precision representation and still maintain the same neural network accuracy, as long as distribution over the weights and activations remains unchanged. 

Adopting this strategy to the entire 4D tensor would yield a very high cropping error, since a single scaling factor $M$ would not be able to single-handedly capture the entire tensor distribution. In this paper we adopt the more general strategy of using a tensor of scaling factors, whose size is adjusted to capture the range of values with higher fidelity. Every tensor of floating point values is divided into two components: one tensor with the same size of the original, composed of low-bit integer values, and another one with a fraction of the size, composed of floating point scaling factors. Each scaling factor is responsible for the scaling of a subgroup of $B$ integer values along its tensor depth dimension, where $B$ is the block size hyperparameter.

The steps taken by \textit{DSConv} are as follows: \textbf{(I)} From a pre-trained network, divide the weight tensor depth-wise into blocks of variable length $B$ and quantize each block; \textbf{(II)} Use the block floating point (BFP) format to quantize the activations, where the block is the same size as the weight tensor; \textbf{(III)} Multiply the integer values of the activations and the weight tensor to maximize inference speed; \textbf{(IV)} Multiply the final values by their respective scales to shift the distribution of the individual blocks to the correct range.

\subsection{Weight Quantization}
\label{subsec:weightquantization}
\begin{figure}[h]
\includegraphics[width=\linewidth]{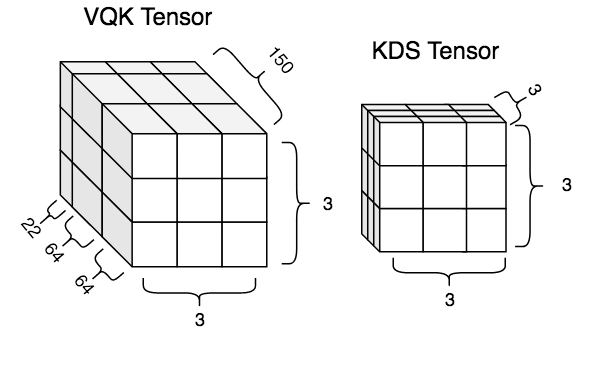}
\caption{Size of VQK and KDS for each weight filter, for the case of $B=64$. This reduces the number of FP MACs from 1350 to 27.}
\label{fig:VQKandKDS}
\end{figure}

We propose a method for quantizing weights that shares one floating-point value for each block of size $B$, along the depth dimension of each weight tensor filter. An example for the resulting sizes for each filter can be seen in Figure \ref{fig:VQKandKDS}. 

Given a weight tensor of size $(C_o, C_i, K_h, K_w)$ and a block size hyperparameter $B$, we first divide the tensor into two components: the \textit{Variable Quantized Kernel} (VQK), which is composed of low-bit values, and is of the same size as the original tensor; and the \textit{Kernel Distribution Shift} (KDS), composed of single precision numbers $\xi$, and of size  $(C_o, \lceil \frac{C_i}{B} \rceil, K_h, K_w)$, where $\lceil x \rceil$ is the ceiling operation.

The $B$ hyperparameter can seamlessly be modified to accommodate for trade-off between floating point arithmetic and fixed point arithmetic, with $B=1$ for pure floating point to $B \geq C_{i}$ for maximum fixed point.

The VQK then holds integer values in 2s complement such that for a specific number of bits $b$ chosen, the weights are in the interval:
\begin{equation}
w_q \in \mathbb{Z}, b \in \mathbb{N} \mid -2^{b-1} \leq w_q \leq 2^{b-1}-1, 
\label{eq:stretch}
\end{equation}

This allows all the operations to be performed using 2s complement arithmetic, as explained in \S\ref{inference}.

By simply changing the normal convolution to \textit{DSConv}, the memory saved per tensor weight is:
\begin{equation}
p = \frac{b}{32}+\frac{\lceil\frac{C_i}{B}\rceil}{C_i}
\label{eq:SavingMemory}
\end{equation}

Equation \ref{eq:SavingMemory} shows that, for large enough values of $B$ and $C_i$, the memory saved is approximately the number of bits $b$ divided by $32$. For illustration purposes, Table \ref{tab:memorySaving} shows the numerical results for realistic values of $C_i$, $B$, $b$ and $p$ for some layers of the GoogLeNet \cite{szegedy2015going} architecture. As it can be seen, significant memory saving can be achieved by only quantizing, with no additional method such as Huffman Coding \cite{huffman1952method}. 

\begin{table}[h]
\small
\centering
\setlength{\tabcolsep}{3pt}
\begin{tabular}{ccccc}
 & Channel ($C_i$) & Block ($B$) & Bit ($b$) & Saving ($p$) \\ \hline
Inception (4a) & 128 & 64 & 4 & 14.1\% \\ \hline
Inception (4a) & 128 & 128 & 4 & 13.3\% \\ \hline
Inception (4a) & 128 & 32 & 3 & 12.5\% \\ \hline
Inception (4c) & 256 & 128 & 3 & 10.2\%
\end{tabular}
\caption{Memory savings by quantizing only.}
\label{tab:memorySaving}
\end{table}

\begin{figure*}[ht]
\centering
\includegraphics[width=0.9\linewidth]{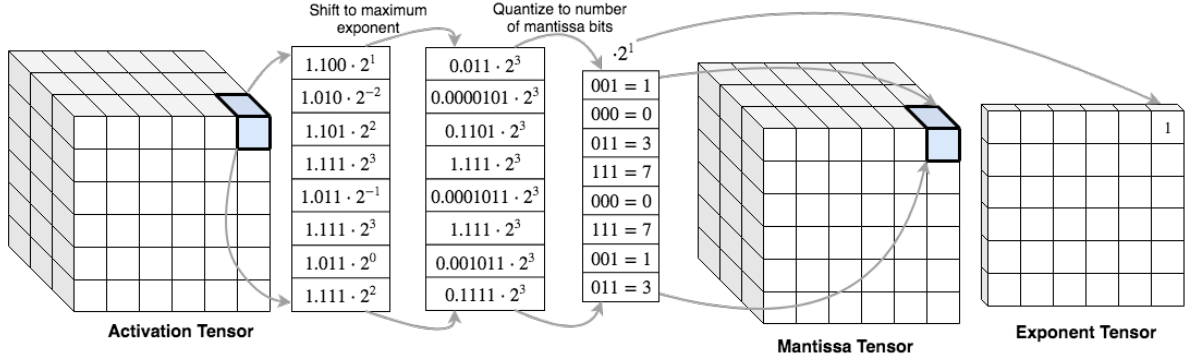}
\caption{Example of quantizing activation. This is the specific case where the mantissa bit was set to 3 and the block hyperparameter was set to 8. Note that \textonehalf{} LSB rounding performed when cropping. Note that this is performed after the ReLU layer, which means that all values are unsigned positive.}
\label{fig:QuantizingActivation}
\end{figure*}

Given a known pre-trained model, the weights of each block are stretched and rounded to fit in the interval in Equation \ref{eq:stretch}, and they are stored in the VQK. Next, we explored two possible methods to calculate the KDS values: (i) minimizing the KL-Divergence, which seeks to find the minimum loss of information between the distribution of the original weights and the kernel distribution shifter and emphasizes the idea that the resulting VQK, after being shifted, should have a similar distribution to the original weights; or (ii) minimizing the L2 norm (Euclidean distance), which can have the interpretation that parameters should be the closest to the optimum value of the original network.

To minimise the KL-Divergence we first take the softmax values of both the shifted VQK and the original distributions:
\begin{equation}
T_j = \frac{e^{w_j}}{\sum_i{e^{w_i}}}, \ I_j = \frac{e^{\hat{\xi}\cdot w_{q_j}}}{\sum_i e^{\hat{\xi}\cdot w_{q_i}}}
\end{equation}

We then use gradient descent to minimize the following for each slice:
\begin{equation}
\xi = \underset{\hat{\xi}}{min} \sum_j T_j \log\left(\frac{T_j}{I_j}\right),\ \ \ \forall \ (1,B,1,1) \ \text{slices}
\end{equation}
where $\xi$ is the KDS value for that block.

The other method minimises the following L2 norm for each slice:
\begin{equation}
\xi = \underset{\hat{\xi}}{min} \sum_{i=0}^{B-1} (w_{q_i}\  \hat{\xi} - w_i)^2
\end{equation}
which has the closed form solution:
\begin{equation}
\therefore\ \  \xi = \frac{\sum_{i=0}^{B-1}w_i w_{q_i}}{\sum_{i=0}^{B-1} w_{q_i}^2},\ \ \ \ \ \ \ \forall \ (1, B, 1, 1)\ \text{slices}
\end{equation}

In practice, both strategies produced approximately equal values. We performed all the experiments using the L2 norm approach, since it has a closed form solution.

Algorithm \ref{alg:weightinit} summarizes the process of initializing both the VQK and the KDS given a pre-trained network model.

\begin{algorithm}
\caption{Weight Initialization}
 \hspace*{\algorithmicindent} \textbf{Input} bit-length $b$, pre-trained weights $\mathbf{w}$, block size $B$ 
\begin{algorithmic}[1]
\Procedure{quantize}{}

\State $\textit{m} \gets 2^{b-1}-1$
\ForAll{Block $B$}:
\State $w_m \gets \argmax_{\mathbf{w}}(|w|)$
\State $\textit{s} \gets m/w_m$
\ForAll{$w_i$ in $B$}:
\State $w_q \gets \round(w_i \cdot s)$
\EndFor
\State $\xi = \sum_{i=0}^{B-1}w_i w_{q_i}/\sum_{i=0}^{B-1} w_{q_i}^2$
\EndFor
\Return $\xi$, $\mathbf{w_q}$ 
\EndProcedure
\end{algorithmic}
\label{alg:weightinit}
\end{algorithm}

\subsection{Activation Quantization}	
\label{subsec:activationquantization}
Our approach aims to achieve good performance in the absence of any training data. This means that we have \textbf{no prior knowledge} of what values or distribution the activation maps will have. Therefore, this quantization cannot be data-driven. Instead, we used an approach inspired by the block floating point (BFP) method of \cite{Wilkinson:1994:REA:1096474, song2018computation, drumond2018training, 8344479, fowers2018configurable}.

Figure \ref{fig:QuantizingActivation} shows our activation quantization approach. For a given mantissa width, the \textit{activation tensor} is divided into blocks, and, for each block, we find the maximum exponent. The mantissa values of all the other activations in the block is shifted such that they match the maximum exponent, which is then cropped (using \textonehalf{} LSB rounding) to match the number of specified bits. This results in two tensors: a \textit{mantissa tensor}, which has the same shape as the original tensor, but populated with $b$ bits; and an \textit{exponent tensor}, which has size $(C_o, \lceil\frac{C_i}{B}\rceil, H, W)$.   

We call this a BFP approach because we are essentially ``sharing" the exponent for each block of size $B$. This allows for a control over how coarse the quantization is, and how much cropping error we are willing to accept to get the lowest bit-length for the mantissa tensor. 

This approach has the added benefit of allowing low-bit integer operations between the weights and activations, as we show in \S\ref{inference}. Therefore, the trade-off between efficiency and speed of computation is as follows: the higher the value of $B$, the bigger the cropping error will be, but the exponent tensor and the KDS will be shallower. This is a different trade-off to the number of bits $b$, which adds more computational complexity and memory to the mantissa tensor. The values of $b$ and $B$ are then inversely proportional to each other and counter-balance each other's positive and negative effects. The goal then becomes to get the most accuracy with the lowest number of mantissa bits $b$ and largest value for the block size $B$.

\subsection{Inference}
\label{inference}
\begin{figure*}[ht]
	\centering
	\includegraphics[width=0.75\linewidth]{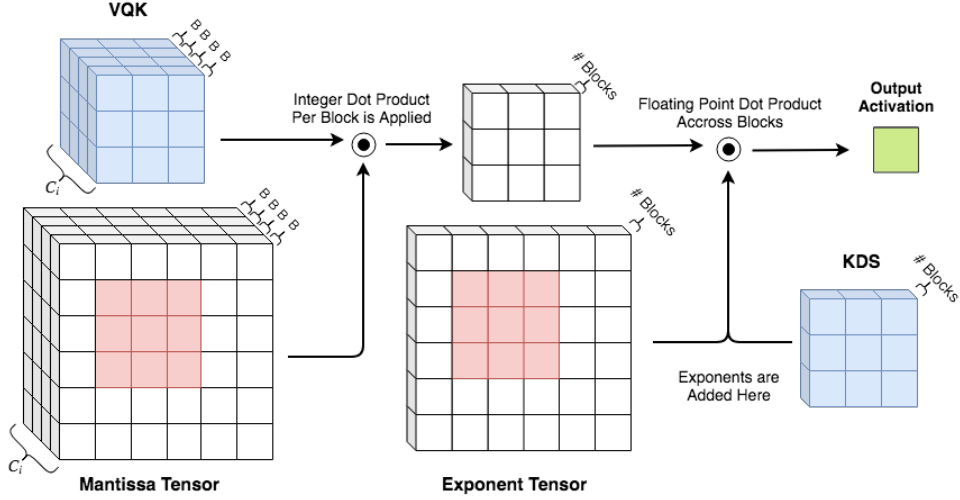}
	\caption{Example of convolution being performed, with VQK tensor (in blue) multiplying one section of the Mantissa Tensor (in red). Each block of the VQK performs a dot product with each block of the Mantissa tensor. The result is a tensor with depth equal to the number of blocks depth-wise. The Exponent Tensor performs addition of the exponent value of the KDS, and the result is multiplied by the result of the dot product of the VQK and the Mantissa tensor, and that becomes the final output activation. This is performed for every filter.}
	\label{fig:InferencePath}
\end{figure*}	
During inference, the hardware can take advantage of the fact that the VQK and the mantissa tensors are low-bit integer values, which allows it to save time performing integer operations rather than floating point operations. The data path is illustrated in Figure \ref{fig:InferencePath}.

First, each of the blocks of the VQK and the mantissa tensor are dot producted, resulting in one value each. All of these operations can be conducted in low-bit fixed point arithmetic, which saves significant processing time. At the end of the block multiplications, the result is a tensor of the same size as both the exponent tensor and the KDS.

The exponent tensor is merged with the KDS tensor by adding its value to the exponent of the KDS tensor values. This results in a tensor of the same size of floating point numbers. Finally, this tensor multiplies the result of the product of the VQK and the KDS, and yields a single floating point number as the output activation.

Notice that the inference is as highly parallelizable as a standard convolution, but instead of performing most of the multiplications using floating point arithmetic, the majority can be substituted by integer multiplications, saving energy, bandwidth and computation time. 

This also means that, for each weight and activation multiplication, the number of blocks is proportional to the number of total floating point MAC operations, and the size of the tensor itself gives the number of INT MAC operations. 

\noindent\textbf{Batch Normalization Folding.} Similar to \cite{jacob2018quantization}, we perform ``folding" of the Batch Normalization (BN) \cite{ioffe2015batch} parameters in models that have them. Since batch normalization has been shown to  improve training (see \cite{ioffe2015batch}), we keep it during the training phase and only fold it for inference.

When folding the BN parameters, we do so with the KDS, since they are unique per channel and use FP32 values. We perform the folding using the equations:
\begin{equation}
\mathbf{\xi}_{fold} = \frac{\mathbf{\xi} \gamma}{\sqrt[]{\sigma_B^2 + \epsilon}}
\end{equation}
\begin{equation}
b_{fold} = \beta - \frac{\gamma\mu_b}{\sqrt[]{\sigma_B^2 + \epsilon}}
\end{equation}
where the parameters $\gamma$, $\sigma$, $\epsilon$, $\beta$ and $\mu$ are as defined in \cite{ioffe2015batch}, $\xi$ is the KDS tensor and $b_{fold}$ is the resulting bias of the \textit{DSConv}.

\subsection{Distillation for Unlabelled Data Adaptation}
\label{subsec:trainingprocedure}
It is often the case that unlabelled data may be available, as shown by the vast array of unsupervised learning methods available. For this specific scenario, we adopt a strategy similar to \cite{hinton2015distilling}. We use the distillation loss without labelled data to try to regress the FP32 model to the quantized one, by using the FP32 logits as the target, and minimizing the loss for regression. 

We create a ``shadow" model which holds single-precision numbers. Before each inference, this model is quantized to the VQK and KDS, inference is performed and the gradients are calculated. During the update phase, the gradients are accumulated as single-precision numbers, and the method is performed until convergence. 

Quantizing the activation maps after each inference would cause the gradients to be zero everywhere. To avoid this problem, we use the Straight-Through Estimator (STE) \cite{bengio2013estimating, yin2018understanding}, to calculate the backwards gradient. Particularly, we use the ReLU STE since it was shown in \cite{yin2018understanding} that it gives a better accuracy than using the Identity STE for deeper networks. The gradient is then also accumulated in a ``shadow" FP32 model, which is quantized after each batch iteration.

We use the ADAM Optimizer \cite{kingma2014adam}, with initial learning rate $10^{-5}$ and after the loss plateaus, this rate is changed to $10^{-6}$. All other hyperparameters and data augmentation details follow their respective original papers.

We use 960 images (30 batches of 32) from the validation dataset for the distillation, and we test the accuracy using the rest of the images (49,040 images in total).

\section{Experiments and Results}
\label{sec:experiments}
We tested our method on various neural network architectures: ResNet \cite{He_2016}, AlexNet \cite{krizhevsky2012imagenet}, GoogleNet \cite{szegedy2015going}, and DenseNet \cite{Huang_2017}. We benchmarked our results on the ImageNet dataset \cite{deng2009imagenet} (more specifically ILSVRC2012), which has 1.2M images in the training set and 50k images in the validation set. The results reported use images drawn from the validation set. We tested our algorithm for all the tasks indicated in the introduction. This section continues as follows: \S\ref{subsec:computationLoad} finds the theoretical computational saving for DSConv; \S\ref{subsec:beforeCal} shows the results without training or adaptation; \S\ref{subsec:adaptation} shows the accuracies when the model is adapted with unlabelled data; and \S\ref{subsec:afterTrainign}, for comparison with previous methods, shows the results for the retraining performed in DSConv using labelled data.

\subsection{Theoretical Computational Load on Block Size}
\label{subsec:computationLoad}
Computational load is traditionally reported as a function of number of MAC operations needed in order to complete the algorithm. We note two caveats: integer MACs are far less complex than FP MACs and, when supported by a hardware implementation, can be run orders of magnitudes faster than FP operations \cite{krishnamoorthi2018quantizing}; our method also relies on the ability to create the mantissa tensor and the exponent tensor dynamically (the VQK and the KDS are created statically, so they are not considered here). This requires \code{MAX}, \code{SHIFT} and \code{MASK} operations. These can be implemented efficiently in custom hardware with few clock cycles. Therefore we will focus on the comparison between number of INT vs FP operations to assess the advantage of using this method.

In order for our method to be faster than normal convolution, the time spent to perform the INT operations must be less than the time spent on the FP32 operations. This difference is a function of the block size and on the channel parameter. Equation \ref{eq:intvsfp} shows the relation between the time for an INT operation and the time for an FP operation:
\begin{equation}
T_{int} \leq T_{FP} \frac{C_i - \lceil \frac{C_i}{B} \rceil}{C_i(1 + \eta)}
\label{eq:intvsfp}
\end{equation}

The values $T_{int}$ and $T_{FP}$ capture the amount of time needed to perform an INT and an FP operation, respectively. The parameter $\eta$ is an ``ideality" parameter that represents the overall overhead in the \code{MAX}, \code{SHIFT} and \code{MASK} operations to perform \textit{DSConv} in comparison to the normal convolution operator.

Also notice that, if $C_i$ is divisible by $B$ (which is often the case), then Equation \ref{eq:intvsfp} becomes independent of the channel size and simplifies to:
\begin{equation}
T_{int} \leq T_{FP} \frac{1 - \frac{1}{B}}{1 + \eta}, \ \ \text{if} \ \ B \mid C_i
\label{eq:intvsfp2}
\end{equation}

Table \ref{tb:blockvsration} shows the ratio $1 - \frac{1}{B}$ for the most common block sizes experimented, when $\eta=0$. As can be seen, if the time to compute an INT value is less than $0.75$ of the time to compute a floating point operation, then all block sizes bigger than 4 will be faster than the normal convolution. This is often likely to be the case. For example, in modern CPUs and in some GPUs, 8-bit operations can be up to 10x faster than FP32 operations \cite{krishnamoorthi2018quantizing}, and lower bit operations can potentially be even faster in custom hardware such as FPGAs. In custom software, operations in less than 8-bit are also often faster.

\begin{table}[h]
\centering
	\begin{tabular}{@{}c|c@{ }@{ }@{ }c@{ }@{ }@{ }c@{ }@{ }@{ }c@{ }@{ }@{ }c@{ }@{ }@{ }c@{ }@{ }@{ }c@{}}
		\hline
		\textbf{Block} & 4 & 8 & 16 & 32 & 64 & 128 \\
		\textbf{Ratio} & 0.750 & 0.875 & 0.938 & 0.969 & 0.984 & 0.992 \\
		\hline
	\end{tabular}
	\caption{Relationship of Block Size and speed ratio needed.}
	\label{tb:blockvsration}
\end{table}

The block size B imposes a limit on how much faster \textit{DSConv} can be over traditional convolution operators. Naturally, \textit{DSConv} can be up to $min(C_i, B)$ times faster than the traditional convolution, since it has $min(C_i, B)$ times less floating point operations than a normal convolution. For example, for block sizes of 128 to 256 and channel sizes of more than 256, \textit{DSConv} can be up to two orders of magnitude faster than a normal convolution.

\subsection{Accuracy Before Retraining or Adaptation}
\label{subsec:beforeCal}
Our method is designed to produce accurate results even when training data is not available, by quantizing from a pre-trained network. 

\begin{table}[h]
\small
\centering
\begin{tabular}{@{}c@{ }@{ }c@{ }|cl|cl|cl@{}}
\textbf{} & \textbf{} & \multicolumn{6}{c}{\textbf{Accuracy (\% @ Top1 and Top5)}} \\ \hline
\textbf{Block} & \textbf{Bit (W/A)} & \multicolumn{2}{c}{\textbf{ResNet50}} & \multicolumn{2}{c}{\textbf{ResNet34}} & \multicolumn{2}{c}{\textbf{ResNet18}} \\ \hline
- & 32 / 32 & 76.1 & \multicolumn{1}{c|}{92.9} & 73.3 & \multicolumn{1}{c|}{91.4} & 69.8 & \multicolumn{1}{c}{89.0} \\ \hline
256&8 / 32& 76.1 & 92.9 & 73.3 & 91.4 & 69.7 & 89.0 \\ \hline
128&6 / 32& 75.9 & 92.8 & 73.2 & 91.4 & 69.5 & 89.0 \\ \hline
16&4 / 32 & 75.1 & 92.3 & 72.6 & 91.0 & 67.7 & 87.8 \\ \hline
4&2 / 32 & 65.1 & 86.2 & 66.8 & 87.6 & 59.1 & 81.7 \\ \hline
128&8 / 8& 76.1 & 92.9 & 73.3 & 91.4 & 69.7 & 89.1 \\ \hline
64&6 / 6& 75.9 & 92.8 & 73.2 & 91.4 & 69.6 & 89.0 \\ \hline
64&5 / 5 & 75.4 & 92.6 & 72.7 & 91.0 & 68.9 & 88.5 \\ \hline
16&4 / 4 & 74.8 & 92.1 & 72.3 & 90.8 & 67.3 & 87.7 \\ \hline
\end{tabular}

\bigskip

\begin{tabular}{@{}c@{ }@{ }c@{ }|cl|cl|cl@{}}
\textbf{} & \textbf{} & \multicolumn{6}{c}{\textbf{Accuracy (\% @ Top1 and Top5)}} \\ \hline
\textbf{Block} & \textbf{Bit (W/A)} & \multicolumn{2}{c}{\textbf{GoogLeNet}} & \multicolumn{2}{c}{\textbf{VGG19}} & \multicolumn{2}{c}{\textbf{Dense121}} \\ \hline
- & 32 / 32 & 67.6 & \multicolumn{1}{c|}{88.3} & 72.4 & \multicolumn{1}{c|}{90.9} & 74.4 & \multicolumn{1}{c}{92.0} \\ \hline
256&8 / 32& 67.6 & 88.3 & 72.3 & 90.9 & 74.4 & 92.0 \\ \hline
128&6 / 32& 67.1 & 88.0 & 72.4 & 90.9 & 74.2 & 91.8 \\ \hline
16&4 / 32 & 63.3 & 85.4 & 72.1 & 90.7 & 72.9 & 91.2 \\ \hline
4&2 / 32 & 27.6 & 51.1 & 69.4 & 89.0 & 62.7 & 84.4 \\ \hline
128&8 / 8& 67.6 & 88.3 & 72.3 & 90.9 & 74.4 & 92.0 \\ \hline
128& 6 / 6& 67.1 & 88.0 & 72.4 & 90.8 & 74.2 & 91.8 \\ \hline
64&5 / 5 & 65.5 & 86.8 & 72.3 & 90.8 & 73.8 & 91.6 \\ \hline
16&4 / 8 & 63.3 & 85.4 & 72.1 & 90.7 & 72.9 & 91.2 \\ \hline
\end{tabular}

\caption{Accuracy of Fast Inference and Compression without data as a function of Bit width (in Weights and Activation) and Block Size.}
\label{tb:ResultsWithoutData}
\end{table}

The second and fifth rows of Table \ref{tb:ResultsWithoutData} show that for both the compression and fast inference problems, no loss of accuracy can be achieved with 8-bit networks even with very high block sizes, as already demonstrated by previous papers and real-life applications \cite{krishnamoorthi2018quantizing, gysel2018ristretto}.  The results also shows that compression down to 4-bits (which in convolutions with channel size input of 256 would yield a \textbf{5x} compression rate) results in an accuracy drop of only 1\% to 2\% depending on the architecture. It can also be seen that very low-bit quantizations become noticeably unstable, varying greatly with architecture. At the extreme, using 2-bits, losses vary by as much as -40\% for GoogLeNet and only -11\% for ResNet50. 

The last four rows show the results for the fast inference problem. Also as known in previous research papers \cite{krishnamoorthi2018quantizing, gysel2018ristretto}, models of 8/8 bits lose only around 0.1\% accuracy. For models of 5/5 and 4/4, we get a drop of 1\% to 3\% in accuracy. To our knowledge, this is the smallest bit-width for fast inference that has been reported when models are neither retrained nor adapted.

The variance with respect to architecture suggests that quantization for 5 or less bits is unstable. However, even for fast-inference with 8-bit accuracy, it can achieve stable and satisfactory results within 1\% of the full precision model. 

\noindent\textbf{Accuracy with respect to Block Size}
Table \ref{tab:AccuracyWithoutDataWRTBlockSize} shows the accuracy with respect to block size. The table shows the results of quantizing the weights only, where the number in parenthesis represents the bit-width of the weights. Naturally, this represents a trade-off between memory and computational load against precision of the network.
\begin{table}[h]
\centering
\small
\begin{tabular}{@{}c@{ }ccccccc@{}}
\textbf{Block} & \textbf{256} & \textbf{128} & \textbf{64} & \textbf{32} & \textbf{16} & \textbf{8} & \textbf{4} \\ \hline
\textbf{ResNet50 (4)} & 73.0 & 73.5 & 73.8 & 74.7 & 75.1 & 75.4 & 75.6 \\ \hline
\textbf{ResNet50 (3)} & 44.6 & 51.9 & 59.6 & 67.4 & 69.6 & 73.6 & 74.7 \\ \hline
\textbf{ResNet34 (4)} & 70.8 & 70.8 & 71.5 & 71.9 & 72.6 & 72.8 & 72.9 \\ \hline
\textbf{ResNet34 (3)} & 59.5 & 60.4 & 63.6 & 66.8 & 69.2 & 70.6 & 71.6 \\ \hline
\textbf{GoogLeNet (4)} & 52.5 & 57.0 & 59.1 & 61.7 & 63.3 & 65.6 & 66.5 \\ \hline
\textbf{GoogLeNet (3)} & 5.7 & 22.4 & 37.6 & 40.3 & 49.2 & 56.8 & 62.5 \\ \hline
\textbf{VGG19 (3)} & 67.6 & 68.6 & 69.5 & 70.4 & 71.1 & 71.6 & 71.8 \\ \hline
\textbf{VGG19 (2)} & 11.3 & 21.8 & 38.1 & 55.5 & 63.1 & 67.5 & 69.4 \\
\end{tabular}
\caption{Accuracy with respect to block size for the compression case with no data available.}
\label{tab:AccuracyWithoutDataWRTBlockSize}
\end{table}
The largest discrepancy in accuracy can be seen in models that use 3 or 2 bit weights. For example, the GoogLeNet model with 3-bits improves its Top1 accuracy from 5.7\% to 56.8\% when changing from a block-size of 256 to 8. 

When using 4-bit quantization schemes, a decrease in the block size achieves accuracy levels that are within 1\% to 2\% of the full precision network. This is the case for example for most networks with block sizes of 16 to 32. 

\subsection{Accuracy Adapted using Unlabelled Data}
\label{subsec:adaptation}
The results when adapting our network with extra unlabelled data are reported in Table \ref{tab:adaptation}. For each Block-Bit configuration, two results are reported: on the top we show the result before adaptation and the bottom (in bold) the result after adaptation using unlabelled data. This strategy increases inference accuracy using 4-bits only for the weights and for the activations to within 2\% of the FP32 precision of the network, even for the extreme cases of using 128 as the block size.

For 3-bits, even though we recover up to 30\% accuracy, there is still a considerable gap between the low-bit accuracies and the full precision ones. For ResNet50 this gap is of 6\% whereas for GoogLeNet it can reach 10\%.

\begin{table}[h]
\centering
\small
\setlength{\tabcolsep}{3pt}
\begin{tabular}{@{}c@{}c@{}c@{}c@{}c@{}c@{}c@{}c@{}c@{}c@{}}
 & \multicolumn{1}{c|}{} & \multicolumn{8}{c}{\textbf{Accuracy (\% @ Top1 and Top5)}} \\ \hline
\textbf{Block} & \multicolumn{1}{c|}{\textbf{\begin{tabular}[c]{@{}c@{}}Bit \\ (W/A)\end{tabular}}} & \multicolumn{2}{c}{\textbf{ResNet50}} & \multicolumn{2}{c}{\textbf{ResNet34}} & \multicolumn{2}{c}{\textbf{ResNet18}} & \multicolumn{2}{c}{\textbf{GoogLeNet}} \\ \hline
\multirow{2}{*}{32} & \multicolumn{1}{c|}{\multirow{2}{*}{4 / 4}} & 74.1 & \multicolumn{1}{c|}{91.8} & 71.3 & \multicolumn{1}{c|}{90.2} & 66.4 & \multicolumn{1}{c|}{87.1} & 61.2 & 81.9 \\
 & \multicolumn{1}{c|}{} & \textbf{74.8} & \multicolumn{1}{c|}{\textbf{92.1}} & \textbf{71.8} & \multicolumn{1}{c|}{\textbf{90.6}} & \textbf{68.3} & \multicolumn{1}{c|}{\textbf{88.1}} & \textbf{66.1} & \textbf{87.2} \\ \hline
\multirow{2}{*}{64} & \multicolumn{1}{c|}{\multirow{2}{*}{4 / 4}} & \multicolumn{1}{l}{73.0} & \multicolumn{1}{l|}{89.8} & \multicolumn{1}{l}{70.9} & \multicolumn{1}{l|}{88.4} & \multicolumn{1}{l}{66.1} & \multicolumn{1}{l|}{85.1} & \multicolumn{1}{l}{58.4} & \multicolumn{1}{c}{79.3} \\
 & \multicolumn{1}{c|}{} & \multicolumn{1}{l}{\textbf{74.8}} & \multicolumn{1}{l|}{\textbf{92.1}} & \multicolumn{1}{l}{\textbf{71.8}} & \multicolumn{1}{l|}{\textbf{90.6}} & \multicolumn{1}{l}{\textbf{68.4}} & \multicolumn{1}{l|}{\textbf{88.1}} & \multicolumn{1}{l}{\textbf{65.2}} & \multicolumn{1}{c}{\textbf{86.8}} \\ \hline
\multirow{2}{*}{128} & \multicolumn{1}{c|}{\multirow{2}{*}{4 / 4}} & \multicolumn{1}{l}{72.6} & \multicolumn{1}{l|}{89.6} & \multicolumn{1}{l}{70.2} & \multicolumn{1}{l|}{87.9} & \multicolumn{1}{l}{65.8} & \multicolumn{1}{l|}{84.8} & \multicolumn{1}{l}{55.8} & \multicolumn{1}{c}{77.3} \\
 & \multicolumn{1}{c|}{} & \multicolumn{1}{l}{\textbf{74.2}} & \multicolumn{1}{l|}{\textbf{92.0}} & \multicolumn{1}{l}{\textbf{71.3}} & \multicolumn{1}{l|}{\textbf{90.5}} & \multicolumn{1}{l}{\textbf{67.5}} & \multicolumn{1}{l|}{\textbf{87.8}} & \multicolumn{1}{l}{\textbf{64.7}} & \multicolumn{1}{c}{\textbf{86.3}} \\ \hline
\multirow{2}{*}{32} & \multicolumn{1}{c|}{\multirow{2}{*}{3 / 3}} & 63.3 & \multicolumn{1}{c|}{85.0} & 63.2 & \multicolumn{1}{c|}{85.0} & 55.3 & \multicolumn{1}{c|}{78.4} & 34.5 & 60.5 \\
 & \multicolumn{1}{c|}{} & \textbf{72.6} & \multicolumn{1}{c|}{\textbf{91.1}} & \textbf{69.6} & \multicolumn{1}{c|}{\textbf{89.4}} & \textbf{66.8} & \multicolumn{1}{c|}{\textbf{87.5}} & \textbf{60.0} & \textbf{83.3} \\ \hline
\multirow{2}{*}{64} & \multicolumn{1}{c|}{\multirow{2}{*}{3 / 3}} & 54.4 & \multicolumn{1}{c|}{77.9} & 58.1 & \multicolumn{1}{c|}{81.3} & 30.1 & \multicolumn{1}{c|}{51.6} & 29.3 & 53.9 \\
 & \multicolumn{1}{c|}{} & \textbf{71.5} & \multicolumn{1}{c|}{\textbf{90.4}} & \textbf{69.1} & \multicolumn{1}{c|}{\textbf{89.3}} & \textbf{65.8} & \multicolumn{1}{c|}{\textbf{87.0}} & \textbf{56.7} & \textbf{81.0} \\ \hline
 &  &  &  &  &  &  &  &  & 
\end{tabular}
\caption{Results of adaptation for a variety of architectures for the case where an adaptation dataset is provided.}
\label{tab:adaptation}
\end{table}

Table \ref{tab:comparsionAdaptation} shows results of recent papers (introduced in the literature review) that use calibration. This is similar in spirit to our adaptation stage, in that both approaches use only unlabelled data. The notable exception is that we use distillation to convert a full-precision model to a low-precision model, whereas the other approaches generally calibrate just the optimal clipping strategy. It can be seen that, even using a big block size of 64, we achieve better performance. To our knowledge, this is the best result achieved for fast inference using only adaptation data.

\begin{table}[h]
\small
\centering
\setlength{\tabcolsep}{2pt}
\begin{tabular}{ccccccc}
 &  &  & VGG16 & AlexNet & ResNet18 & ResNet50 \\ \cline{2-7} 
 & W & A & Top1 & Top1 & Top1 & Top1 \\ \cline{2-7} 
Naive \cite{banner2018posttraining}& 4 & \multicolumn{1}{c|}{8} & 29.0\% & 1.8\% & 0.8\% & 0.4\% \\
CW \cite{banner2018posttraining, krishnamoorthi2018quantizing}& 4 & \multicolumn{1}{c|}{8} & 70.2\% & 52.9\% & 59.3\% & 72.4\% \\
K+B \cite{banner2018posttraining}& 4 & \multicolumn{1}{c|}{8} & 70.0\% & 54.7\% & 67.0\% & 74.2\% \\
OCS+MSE \cite{zhao2019improving} & 5 & \multicolumn{1}{c|}{8} & - & - & - & 73.4\% \\
\textbf{Ours NA (16)} & 4 & \multicolumn{1}{c|}{8} & \textbf{71.3\%} & \textbf{55.9\%} & \textbf{67.6\%} & \textbf{75.1\%} \\
\textbf{Ours NA (32)} & 4 & \multicolumn{1}{c|}{8} & \textbf{71.2\%} & \textbf{55.4\%} & \textbf{66.7\%} & \textbf{74.7\%} \\
Naive \cite{banner2018posttraining} & 8 & \multicolumn{1}{c|}{4} & 53.9\% & 41.6\% & 53.2\% & 52.7\% \\
KLD \cite{banner2018posttraining, TensorRT}& 8 & \multicolumn{1}{c|}{4} & 67.0\% & 49.6\% & 65.1\% & 70.8\% \\
ACIQ \cite{banner2018posttraining} & 8 & \multicolumn{1}{c|}{4} & 70.5\% & 55.2\% & 68.9\% & 74.8\% \\
\textbf{Ours NA (16)} & 8 & \multicolumn{1}{c|}{4} & \textbf{71.5\%} & \textbf{56.4\%} & \textbf{69.6\%} & \textbf{75.7\%} \\
\textbf{Ours NA (32)} & 8 & \multicolumn{1}{c|}{4} & \textbf{71.5\%} & \textbf{56.4\%} & \textbf{69.6\%} & \textbf{75.6\%} \\
Naive \cite{banner2018posttraining} & 4 & \multicolumn{1}{c|}{4} & 23.7\% & 1.8\% & 0.6\% & 0.4\% \\
ACIQ \cite{banner2018posttraining} & 4 & \multicolumn{1}{c|}{4} & 68.9\% & 53.0\% & 65.3\% & 72.6\% \\
OMSE+O \cite{choukroun2019lowbit}& 4 & \multicolumn{1}{c|}{4} & - & 54.5\% & 67.4\% & 72.6\% \\
\textbf{Our (64)} & 4 & \multicolumn{1}{c|}{4} & \textbf{71.1\%} & \textbf{55.8\%} & \textbf{68.4\%} & \textbf{74.8\%}
\end{tabular}
\caption{Adaptation of our method vs previous papers. The Naive method refers to simple clipping. CW is Channel-Wise quantization adopted in \cite{wu2018training, krishnamoorthi2018quantizing}. K+B is the K-Means + Bias method of \cite{banner2018posttraining}. KLD is the KL-Divergence method first proposed in \cite{TensorRT}. OMSE+O is the OMSE + offset method of \cite{choukroun2019lowbit}. ``Ours NA" refer to our method with No Adaptation}
\label{tab:comparsionAdaptation}
\end{table}

\subsection{Accuracy After Labelled Data Retraining}
\label{subsec:afterTrainign}
\begin{table}[]
\small
\centering
\begin{tabular}{@{}c@{ }c@{ }c@{ }c@{ }c@{ }l@{ }c@{ }c@{ }c@{ }c@{ }c@{}}
\multicolumn{5}{c}{ResNet 18} &  & \multicolumn{5}{c}{ResNet34} \\ \cline{1-5} \cline{7-11} 
\multicolumn{1}{l}{} & W & A & Top1 & Top5 &  &  & W & A & Top1 & Top5 \\ \cline{2-5} \cline{8-11} 
FP & 32 & 32 & 69.6 & 89.2 &  & FP & 32 & 32 & 73.3 & 91.3 \\
FP$_{\text{LQ}}$ \cite{Zhang_2018} & 32 & 32 & 70.3 & 89.5 &  & FP$_{\text{LQ}}$ \cite{Zhang_2018} & 32 & 32 & 73.8 & 91.4 \\
BWN \cite{Rastegari_2016}& 1 & 32 & 60.8 & 83.0 &  & \textbf{Ours (32)} & 3 & 32 & \textbf{73.4} & \textbf{90.1} \\
TWN \cite{li2016ternary}& 2 & 32 & 61.8 & 84.2 &  & \textbf{Ours (32)} & 4 & 32 & \textbf{73.6} & \textbf{90.1} \\
TWN \cite{li2016ternary} & 2 & 32 & 65.3 & 86.2 &  & HWGQ \cite{Cai_2017}& 1 & 2 & 64.3 & 85.7 \\
TTQ \cite{zhu2016trained}& 2 & 32 & 66.6 & 87.2 &  & \textbf{Ours (64)} & 1 & 4 & \textbf{68.2} & \textbf{86.8} \\
LQ \cite{Zhang_2018} & 2 & 32 & 68.0 & 88.0 &  & ABC \cite{lin2017towards} & 3 & 3 & 66.7 & 87.4 \\
\textbf{Ours (32)} & 2 & 32 & \textbf{68.7} & \textbf{86.7} &  & ABC \cite{lin2017towards} & 5 & 5 & 68.4 & 88.2 \\
LQ \cite{Zhang_2018}& 3 & 32 & 69.3 & 88.8 &  & \textbf{Ours(16)} & 4 & 4 & \textbf{73.0} & \textbf{89.7} \\
\textbf{Ours (32)} & 3 & 32 & \textbf{69.7} & \textbf{87.5} &  & LQ\cite{Zhang_2018} & 1 & 2 & 66.6 & 86.9 \\
LQ \cite{Zhang_2018}& 4 & 32 & 70.0 & 89.1 &  & LQ \cite{Zhang_2018}& 2 & 2 & 67.8 & 89.1 \\
\textbf{Ours (32)} & 4 & 32 & \textbf{70.0} & \textbf{87.6} &  & LQ \cite{Zhang_2018}& 3 & 3 & 71.9 & 90.2 \\
XNOR \cite{Rastegari_2016} & 1 & 1 & 51.2 & 73.2 &  & \textbf{Ours (16)} & 3 & 3 & \textbf{72.7} & \textbf{89.6} \\
DoReFa \cite{zhou2016dorefa} & 1 & 2 & 53.4 & - &  & \multicolumn{1}{l}{} & \multicolumn{1}{l}{} & \multicolumn{1}{l}{} & \multicolumn{1}{l}{} & \multicolumn{1}{l}{} \\
DoReFa \cite{zhou2016dorefa} & 1 & 4 & 59.2 & - &  & \multicolumn{5}{c}{ResNet50} \\ \cline{7-11} 
\textbf{Ours (32)} & 1 & 4 & \textbf{65.2} & \textbf{86.2} &  &  & W & A & Top1 & Top5 \\ \cline{8-11} 
HWGQ \cite{Cai_2017} & 1 & 2 & 59.6 & 82.2 &  & FP & 32 & 32 & 76.0 & 93.0 \\
ABC \cite{lin2017towards} & 3 & 3 & 61.0 & 83.2 &  & FP$_{\text{LQ}}$ \cite{Zhang_2018} & 32 & 32 & 76.4 & 93.2 \\
ABC \cite{lin2017towards} & 5 & 5 & 65.0 & 85.9 &  & LQ\cite{Zhang_2018} & 2 & 32 & 75.1 & 92.3 \\
\textbf{Ours (128)} & 5 & 5 & \textbf{70.0} & \textbf{89.3} &  & \textbf{Ours (32)} & 2 & 32 & \textbf{75.2} & \textbf{92.6} \\
LQ \cite{Zhang_2018}& 1 & 2 & 62.6 & 84.3 &  & LQ\cite{Zhang_2018} & 4 & 32 & 76.4 & 93.1 \\
LQ\cite{Zhang_2018} & 2 & 2 & 64.9 & 68.2 &  & \textbf{Ours(128)} & 4 & 32 & \textbf{76.4} & \textbf{93.0} \\
LQ \cite{Zhang_2018}& 3 & 3 & 68.2 & 87.9 &  & HWGQ \cite{Cai_2017} & 1 & 2 & 64.6 & 85.9 \\
\textbf{Ours (16)} & 3 & 3 & \textbf{69.2} & \textbf{88.9} &  & ABC \cite{lin2017towards} & 5 & 5 & 70.1 & 89.7 \\
LQ \cite{Zhang_2018}& 4 & 4 & 69.3 & 88.8 &  & LQ\cite{Zhang_2018} & 1 & 2 & 68.7 & 88.4 \\
\textbf{Ours (64)} & 4 & 4 & \textbf{69.8} & \textbf{89.2} &  & LQ\cite{Zhang_2018} & 2 & 2 & 71.5 & 90.3 \\
\multicolumn{1}{l}{} & \multicolumn{1}{l}{} & \multicolumn{1}{l}{} & \multicolumn{1}{l}{} & \multicolumn{1}{l}{} &  & \textbf{Ours(32)} & 2 & 2 & \textbf{72.5} & \textbf{91.2} \\
\multicolumn{5}{c}{DenseNet121} &  & LQ\cite{Zhang_2018} & 3 & 3 & 74.2 & 91.6 \\ \cline{1-5}
 & W & A & Top1 & Top5 &  & \textbf{Ours (32)} & 3 & 3 & \textbf{75.2} & \textbf{92.4} \\ \cline{2-5}
FP & 32 & 32 & 75.0 & 92.3 &  & LQ\cite{Zhang_2018} & 4 & 4 & 75.1 & 92.4 \\
DoReFa \cite{zhou2016dorefa} & 2 & 2 & 67.7 & 88.4 &  & \textbf{Ours (64)} & 4 & 4 & \textbf{76.2} & \textbf{92.9} \\
FP$_{\text{LQ}}$ \cite{Zhang_2018}& 32 & 32 & 75.3 & 92.5 &  & \textbf{Ours (128)} & 4 & 4 & \textbf{76.1} & \textbf{92.8} \\
LQ \cite{Zhang_2018}& 2 & 2 & 69.6 & 89.1 &  & \multicolumn{1}{l}{} & \multicolumn{1}{l}{} & \multicolumn{1}{l}{} & \multicolumn{1}{l}{} & \multicolumn{1}{l}{} \\
FP$_{\text{Ours}}$& 32 & 32 & 74.4 & 92.2 &  & \multicolumn{5}{c}{GoogLeNet} \\ \cline{7-11} 
\textbf{Ours (32)} & 2 & 32 & \textbf{74.0} & \textbf{91.8} &  &  & W & A & Top1 & Top5 \\ \cline{8-11} 
\textbf{Ours (16)} & 2 & 2 & \textbf{72.1} & \textbf{90.6} &  & FP$_{\text{HWGQ}}$ \cite{Cai_2017} & 32 & 32 & 71.4 & 90.5 \\
\multicolumn{1}{l}{} & \multicolumn{1}{l}{} & \multicolumn{1}{l}{} & \multicolumn{1}{l}{} & \multicolumn{1}{l}{} &  & HWGQ \cite{Cai_2017}& 1 & 2 & 63.0 & 84.9 \\
\multicolumn{5}{c}{AlexNet} &  & FP$_{\text{LQ}}$ \cite{Zhang_2018} & 32 & 32 & 72.9 & 91.3 \\ \cline{1-5}
 & W & A & Top1 & Top5 &  & LQ \cite{Zhang_2018} & 1 & 2 & 65.6 & 86.4 \\ \cline{2-5}
FP & 32 & 32 & 57.1 & 80.2 &  & LQ \cite{Zhang_2018} & 2 & 2 & 68.2 & 88.1 \\
TWN \cite{li2016ternary} & 2 & 32 & 54.5 & 76.8 &  & FP$_{\text{Ours}}$ & 32 & 32 & 67.6 & 86.3 \\
FP$_{\text{LQ}}$ \cite{Zhang_2018}& 32 & 32 & 61.8 & 83.5 &  & \textbf{Ours (32)} & 4 & 4 & \textbf{66.3} & \textbf{85.5} \\
LQ\cite{Zhang_2018} & 2 & 32 & 60.5 & 82.7 &  & \textbf{Ours (64)} & 4 & 4 & \textbf{65.7} & \textbf{85.1} \\
\textbf{FP$_{\text{Ours}}$} & 32 & 32 & \textbf{56.6} & \textbf{79.1} &  & \multicolumn{1}{l}{} & \multicolumn{1}{l}{} & \multicolumn{1}{l}{} & \multicolumn{1}{l}{} & \multicolumn{1}{l}{} \\
\textbf{Ours (32)} & 2 & 32 & \textbf{55.0} & \textbf{78.1} &  & \multicolumn{1}{l}{} & \multicolumn{1}{l}{} & \multicolumn{1}{l}{} & \multicolumn{1}{l}{} & \multicolumn{1}{l}{}
\end{tabular}
\caption{Results of retraining for a variety of architectures. Table derived from \cite{Zhang_2018}}
\label{tab:retrainingres}
\end{table}
We also compared \textit{DSConv} with previous methods that retrain/finetune with labelled data (the vast majority in the literature). Training happens similarly to adaptation, but now we use labelled data and use cross-entropy loss in the classification error instead of using logits.

Table \ref{tab:retrainingres} shows the results for ImageNet on a variety of architectures. As many previous papers report different initial FP accuracy for the same architecture, we have also included the initial FP of the single precision results to make an evaluation that takes into account the ``upper limit" of the architecture itself.

From the results, it can be seen that our method can beat the state-of-the-art for a variety of cases, as long as the Block Size is adjusted properly to give more emphasis on accuracy rather than speed.

\textit{DSConv} can beat the state-of-the-art when using bit sizes that are either 4 or 5. In these cases (such as ResNet18 using 5/5, ResNet50 using 4/4 and GoogLeNet using 4/4), we also use a large Block Size, with slightly better than the FP efficiency in ResNet18 when using $B=128$ and bit sizes of 5/5, and $B=64$ for bit sizes of 4/4.

In order to get state-of-the-art results for 3-bits or less, a lower block size is needed. This is shown for DenseNet121 results, which uses bit-width of 2 and Block Size of 16 to get 72.1\% accuracy. Extremely low-bit weights and activations do not work very well because the assumption that lower information loss in quantization corresponds to higher accuracy starts to break down. This is supported by the fact that the state-of-the-art approaches for 1 and 2 bit weights are trained from scratch, which suggests that for these cases, quantizing from a pre-trained network is not ideal.

We also show good results for the compression case. ResNet50 with 4 bit and $B=128$ illustrates that no loss of accuracy is observed, and even using only 2 bits, with accuracy staying within 1\% using $B=32$.

\section{Conclusion}
\label{sec:conclusion}
We presented \textit{DSConv}, which proposes an alternative convolution operator that can achieve state-of-the-art results whilst quantizing models to up to 4-bits in weight and activation without retraining or adaptation.

We showed that our method can achieve state-of-the-art results without retraining in less than 8 bit settings, which makes it possible for fast inference and less power consumption for rapid deployment in custom hardware. By having the advantage of being tunable by the block size hyperparameter and not needing any training data in order to run, we propose that this method is very suitable for acceleration of convolutional neural networks of any kind.

When using unlabelled data and distillation from the FP32 model, we can achieve less than 1\% loss using 4-bit for both weights and activations. Also, as in previous methods, we demonstrate that the assumption that lower information loss in weight quantization correspond to higher inference accuracy breaks down when quantizing to extremely low bits (1, 2 or 3 bits). In these cases, retraining seems inevitable since they are quintessentially different than higher accuracy models.

\section{Acknowledgements}

This research was partially supported by Intel (previously Omnitek), and we thank our colleagues from the Programmable Solutions Group who greatly assisted in this work. Victor Adrian Prisacariu would like to thank the European Commission Project Multiple-actOrs Virtual Empathic CARegiver for the Elder (MoveCare).

\newpage
{\small
\bibliographystyle{ieee_fullname}
\bibliography{egbib}
}

\end{document}